\documentclass{article}

\PassOptionsToPackage{numbers, compress}{natbib}
\usepackage[final,sglblindworkshop]{neurips_2026}
\workshoptitle{TS-LIMITS}

\usepackage[utf8]{inputenc}
\usepackage[T1]{fontenc}
\usepackage{hyperref}
\usepackage{url}
\usepackage{booktabs}
\usepackage{amsfonts}
\usepackage{amsmath}
\usepackage{array}
\usepackage{nicefrac}
\usepackage{microtype}
\usepackage{xcolor}
\usepackage{graphicx}
\usepackage{float}
\usepackage{algorithm}
\usepackage{algorithmic}

\title{TMAML: Temporal Model-Agnostic Meta-Learning for Cold-Start Time Series Forecasting}

\author{%
  Wannes Janssens, Matthias Bogaert, Dirk Van den Poel \\
  Ghent University, Research Group Data Analytics, Ghent, Belgium \\
  FlandersMake@UGent--corelab CVAMO \\
  \texttt{wanjanss.janssens@UGent.be}
}

\begin{document}

\maketitle

\begin{abstract}
Cold-start forecasting, the task of forecasting a time series with little to no historical data, is a common challenge. Addressing it requires approaches that learn quickly from few datapoints and leverage information from related series, typically through static covariates, to generalize well to unseen series. While some global forecasting models can generate cold-start predictions by leveraging information shared across multiple series, they are not optimized for out-of-train-set generalization or adaptation from short histories. In this work, we formulate cold-start forecasting as a few-window learning problem and introduce Temporal Model-Agnostic Meta-Learning (TMAML), which tailors the model-agnostic meta-learning algorithm, originally developed for few-shot adaptation of neural networks, to deep time series forecasting. TMAML constructs meta-tasks as temporally consistent support-query windows and pairs them with a temporal meta-training and meta-testing procedure, yielding forecasting models that are explicitly optimized for cold-start forecasting. We instantiate TMAML on the Temporal Fusion Transformer (TFT) and present an initial empirical analysis of forecast accuracy and calibration across three cold-start scenarios: TMAML consistently outperforms or matches a standard ERM-trained TFT, yields better-calibrated forecasts than naive on two of the three scenarios, but does not consistently outperform naive on probabilistic forecast accuracy.
\end{abstract}


\begin{figure}
    \makebox[\textwidth][c]{\includegraphics[width=1.1\textwidth]{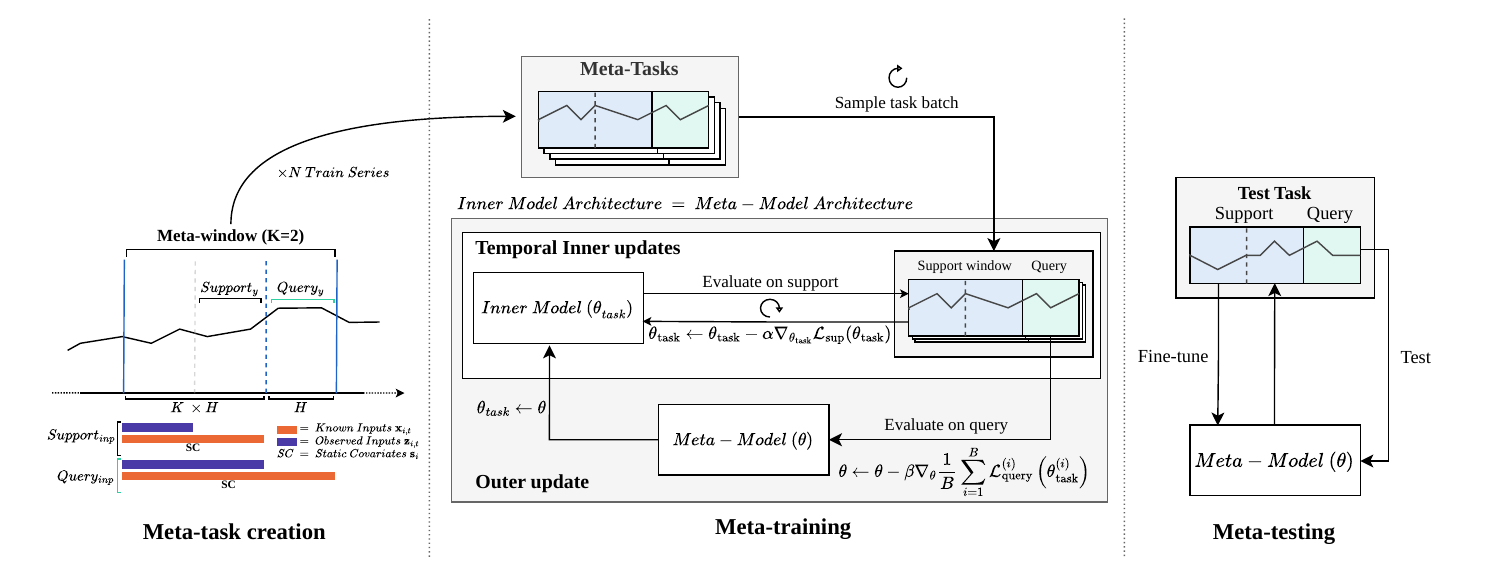}}
    \centering
    \caption{Illustration of the TMAML framework. The meta-window slides across the observed series to create multiple train meta-tasks, each consisting of a support sample and a query sample. The inner model is adapted on the support sample and evaluated on the query sample, whose loss is used to update the meta-model. The figure shows the TMAML process for K=2.}
    \label{fig:tmaml_framework}
\end{figure}

\section{Introduction}

The cold-start problem arises when a time series has little to no historical data available, which makes accurate forecasting particularly challenging. This setting is common in practice, for example in demand forecasting for newly launched products or consumption forecasting for newly connected electricity customers. Such cold-start forecasts carry high business value, yet are traditionally produced by domain experts relying on judgement and heuristics. As new series continue to emerge at scale, this manual approach becomes increasingly impractical, motivating the need for automatic cold-start forecasting methods. Forecasting a cold-start time series typically relies on information available in related series, under the assumption that series sharing similar attributes tend to follow similar temporal patterns. Relatedness is commonly defined through static covariates, such as product family or customer segment, but can also be inferred from auxiliary information such as textual descriptions or images.

Several retrieval-based frameworks put this principle into practice, matching a cold-start series to related series and transferring information from them: through static-feature similarity \citep{chauhan2020time,caserta2025intermittent}, a causal graph over covariates \citep{fatemi2023mitigating}, or textual embeddings of unstructured content \citep{zhou2025context}. What these methods share is that each is non-generic: tailored to a single application, and reliant on manually defined prior knowledge of both how series are related and how to measure that relatedness.

Global forecasting models offer an alternative way of exploiting relatedness between series. Rather than explicitly retrieving related series through defined attributes and similarity measures, they learn shared temporal structure implicitly by training jointly across a collection of related series. Most architectures, however, cannot natively incorporate static covariates or handle variable-length inputs effectively, which limits their applicability to the cold-start setting. A notable exception is the Temporal Fusion Transformer (TFT) \citep{lim2021temporal}, which incorporates static covariates and supports variable-length inputs, and can therefore in principle forecast a cold-start series and refine that forecast as more data becomes available. It is nonetheless trained like any other global model to perform well on average across the series it was trained on, and is never explicitly optimised to generalise to unseen series or to adapt quickly from a short history. This is the gap we address.

Specifically, we make the following contributions. First, we formalise cold-start forecasting as a few-window learning problem, giving it a precise mathematical definition. Second, we propose Temporal-MAML (TMAML), a temporal adaptation of the MAML \citep{finn2017model} framework tailored to cold-start time series forecasting, which introduces two key components: (i) temporal support-query splits that explicitly emulate cold-start conditions, and (ii) temporal inner-loop updates during meta-training and meta-testing. Together, these components yield a meta-learned initialisation that is explicitly optimised for rapid adaptation to a new time series with limited history. Third, we evaluate TMAML-TFT, an instantiation of TMAML on the TFT, against both a baseline TFT trained with standard empirical risk minimisation (ERM) and a naive empirical-quantile baseline, on two cold-start retail demand-forecasting datasets and one cold-start electricity-consumption dataset, all derived from real, existing benchmark datasets, and show that TMAML-TFT matches or outperforms the ERM baseline throughout, yields better-calibrated forecasts than naive on two of the three datasets, but does not consistently surpass naive on probabilistic forecast accuracy (WQL), indicating that TMAML's meta-learning dynamics, and the cold-start setting itself, warrant deeper analysis.

\section{Related Work}
\label{sec:related}

A cold-start forecaster needs two capabilities: it must exploit information from related series, typically via static covariates, since a cold-start series by definition offers little history of its own, and it must handle variable-length input, including the case of no history at all. Classical local models, including ARIMA, Exponential Smoothing \citep{hyndman2018forecasting}, and Theta \citep{assimakopoulos2000theta}, fit a separate set of parameters to each individual series and thus meet neither requirement. Time series foundation models \citep{das2024decoder,ansari2024chronos,liu2025sundial,auer2025tirex}, despite large-scale pretraining, condition forecasts at inference time primarily on a long local historical window and generally have no mechanism for incorporating static, series-level covariates, the opposite of the conditions the cold-start setting offers. 

Global models instead share parameters across the series of one dataset, learning patterns common to a set of related series, and span several architectural families: recurrent networks \citep{rangapuram2018deep,salinas2020deepar,rasul2021autoregressive}, stacked feedforward networks \citep{oreshkin2019n,challu2023nhits}, and transformer-based models \citep{lim2021temporal,nie2022time,wang2024timexer}. Despite this diversity, many cannot handle variable-length input or incorporate static metadata effectively, and are typically trained with ERM, which optimises for accuracy on the series and window lengths seen during training, with no explicit incentive to generalise to unseen series or adapt from a short history. We build on the TFT \citep{lim2021temporal}, which processes static covariates through a dedicated variable-selection and static-enrichment mechanism, and encodes observed history with a recurrent layer followed by attention rather than fixed-size patches, so it accepts an input window of any length, including an empty one. Architecturally, TFT is therefore well suited to cold-start forecasting.

Model-Agnostic Meta-Learning (MAML) \citep{finn2017model} is an optimisation-based meta-learning method that learns a set of initial neural network parameters that can be quickly fine-tuned to a new task using only a few gradient steps, via a bi-level inner/outer-loop optimisation, and has demonstrated strong few-shot performance in domains such as computer vision \citep{finn2017model} and recommendation systems \citep{lee2019melu}. MAML has also been applied to time series problems \citep{jiang2024metatrans,mo2023few,arango2021multimodal}, but in these applications tasks are always defined at the entity or dataset level \citep{wu2025meta}, where each task comprises multiple related series and full historical data is assumed available at test time. We instead define each task as a single time series and target temporal cold-start forecasting, where adaptation must occur from a short initial history rather than across multiple series belonging to the same entity.

\section{Problem Definition}
\label{sec:problemdef}

In this section, we introduce notation and define cold-start forecasting as a few-window learning problem, using the term ``window'' rather than ``shot'' to avoid confusion with the more common few-shot setting, where a task is instead defined as a set of related series.

Consider a set of $I$ entities, such as products or customers, indexed by $i \in \mathcal{I}$. Each entity is associated with static covariates $\mathbf{s}_i$. At each time step $t$, we observe scalar targets $y_{i,t}$, observed inputs $\mathbf{z}_{i,t}$, and known inputs $\mathbf{x}_{i,t}$. Observed inputs are only revealed at time $t$ and are not available in advance, whereas known inputs are predetermined and available for all future time steps.

At a given time $t$, the set of entities can be partitioned into two disjoint subsets $\mathcal{I} = \mathcal{I}^{\text{obs}} \cup \mathcal{I}^{\text{cold}}$. The subset $\mathcal{I}^{\text{obs}}$ contains entities for which historical targets and observed inputs are available, while $\mathcal{I}^{\text{cold}}$ consists of cold-start entities for which no past observations are available at time $t$. For entities $i \in \mathcal{I}^{\text{obs}}$, standard global forecasting models produce predictions of the form
\begin{equation}
y_{i,t+h} = f_{\text{obs}}\!\left(
y_{i,t-k:t}, \mathbf{z}_{i,t-k:t}, \mathbf{x}_{i,t-k:t+h}, \mathbf{s}_i
\right),
\label{eq:standard_forecast}
\end{equation}
where $h \in \{1, \ldots, H\}$ denotes the forecast horizon, $k \leq L$ is the length of the lookback window bounded by a maximum history length $L$, and $f_{\text{obs}}(\cdot)$ is a forecasting model trained using the historical data of entities in $\mathcal{I}^{\text{obs}}$.

We define cold-start forecasting as the task of generating predictions for entities either with no historical data (zero-window forecasting) or a limited number of observed history windows (K-window forecasting), where each window has length $H$. Since cold-start entities initially contain little or no historical information, K-window forecasting proceeds sequentially, starting from the zero-window case and then applying the same $f_{\text{obs}}$ as the history grows one window at a time,
\begin{align*}
\text{0-window:} \quad
& y_{i,t:t+H} = f_{\text{obs}}\!\left( \mathbf{x}_{i,t:t+H}, \mathbf{s}_i \right), \\
\text{$k$-window:} \quad
& y_{i,t+kH:t+(k+1)H} = f_{\text{obs}}\!\left(
y_{i,t:t+kH}, \mathbf{z}_{i,t:t+kH}, \mathbf{x}_{i,t:t+(k+1)H}, \mathbf{s}_i
\right), \ \ k = 1, 2, \ldots,
\end{align*}
until the total available history reaches the maximum lookback window length $L$. An illustration of the temporal cold-start forecasting setting is provided in Figure~\ref{fig:coldstart_problem}, Appendix~\ref{sec:appendix_problemdef}.

\section{Method}
\label{sec:method}

Building on the $K$-window formulation of Section~\ref{sec:problemdef}, we introduce TMAML: a temporal adaptation of MAML that meta-learns an initialization for a forecasting model $f_\theta$ explicitly optimized for this setting. TMAML is model-agnostic: $f_\theta$ can be any forecaster that consumes the inputs of Section~\ref{sec:problemdef}, predicts $H$ steps ahead and is optimized with gradient descent. In our experiments we instantiate it with a TFT. The framework consists of three parts, meta-task creation, meta-training, and meta-testing, illustrated in Figure~\ref{fig:tmaml_framework} for $K=2$.

\subsection{Meta-Task Creation}
\label{sec:metatask}
Each training meta-task is drawn from an observed series as a meta-window: a contiguous span of $(K+1)H$ timesteps of entity $i$'s series, starting at some time $t$ within that series. The first $KH$ timesteps are the support region and the final $H$ timesteps are the query horizon. From this meta-window we build two samples, used for the outer and inner loops of meta-training, respectively. The query sample uses the entire support region as input and the true query horizon as target,
\begin{equation}
y_{i,t+KH:t+(K+1)H} = f_\theta\big(y_{i,t:t+KH}, \mathbf{z}_{i,t:t+KH}, \mathbf{x}_{i,t:t+(K+1)H}, \mathbf{s}_i\big).
\end{equation}
The support region supplies the inner loop's own supervised signal. We hold out the \emph{last} $H$ timesteps of the support region as its target and use everything before that as input: the model adapts by learning to predict the most recent window of the support region from everything observed before it, one window short of the actual query horizon,
\begin{equation}
y_{i,t+(K-1)H:t+KH} = f_\theta\big(y_{i,t:t+(K-1)H}, \mathbf{z}_{i,t:t+(K-1)H}, \mathbf{x}_{i,t:t+KH}, \mathbf{s}_i\big).
\end{equation}
For $K=0$ the support region is empty, so no such target exists and the meta-task reduces to the $0$-window case, i.e.\ zero-shot prediction with no adaptation. Meta-tasks are sampled uniformly at random across series and window positions during training. Validation and testing uses only the final meta-window of each held-out series for a stable loss.

\subsection{Meta-Training}
\label{sec:metatraining}
We meta-train with First-Order MAML (FOMAML), which drops the second-order terms of the original MAML update for computational efficiency. For each sampled meta-task, an inner loop initializes task-specific parameters $\theta_{\text{task}} \leftarrow \theta$ and takes $M$ gradient steps on the support loss,
\begin{equation}
\theta_{\text{task}} \leftarrow \theta_{\text{task}} - \alpha \nabla_{\theta_{\text{task}}} \mathcal{L}_{\text{support}}(\theta_{\text{task}}),
\end{equation}
with inner learning rate $\alpha$. Meta-tasks are grouped into meta-batches of $B$ tasks, sampled entity-diverse so that each meta-batch draws from $B$ distinct entities rather than repeatedly from the same series. An outer loop evaluates the adapted parameters of every task in the meta-batch on their query sample and updates the meta-parameters $\theta$ using the average query loss,
\begin{equation}
\theta \leftarrow \theta - \beta \nabla_{\theta} \frac{1}{B}\sum_{i=1}^{B} \mathcal{L}_{\text{query}}^{(i)}(\theta_{\text{task}}^{(i)}),
\end{equation}
with outer learning rate $\beta$, treating the inner-loop trajectory as fixed rather than differentiating through it, which is the FOMAML approximation. Setting $K=0$ skips the inner loop entirely, so the same objective also covers zero-shot evaluation. The full procedure is summarized in Algorithm~\ref{alg:tmaml} (Appendix~\ref{sec:appendix_algo}).

\subsection{Meta-Testing}
\label{sec:metatesting}
At meta-test time we emulate deployment on a new cold-start series, following the same $K$-window progression as Section~\ref{sec:problemdef}. For a certain value of $k$ we evaluate, the meta-trained parameters $\theta$ are copied to $\theta_{\text{task}}$ and adapted with $M$ gradient steps on the support sample of the $k$-window meta-task, built from the series' actual observed history exactly as in the inner loop of meta-training. This adaptation step is applied identically to the ERM baseline, giving it the same $k$-shot fine-tuning opportunity. The adapted parameters are then used to forecast the true, unseen query, i.e.\ the $H$ steps immediately following the available history. This is repeated independently for all held-out meta-test series and for several values of $k$. The final evaluation metric is averaged across all test series for each $k$.

\section{Experimental Setup}
\label{sec:expsetup}

In an initial set of experiments, we compare a TFT trained with standard ERM, a TFT trained with our TMAML framework, and a naive baseline across three cold-start forecasting scenarios.\footnote{Code is available at \url{https://github.com/Wannes15/TMAML_tslimits}.}

\subsection{Construction of Cold-Start Datasets}
\label{sec:construction}
Due to limited availability of publicly available datasets specifically designed for cold-start forecasting, we constructed three dataset scenarios from existing benchmark time series forecasting datasets, spanning two domains and three temporal frequencies. We emulate two cold-start retail demand-forecasting scenarios, based on the Favorita and M5 datasets \citep{favorita-grocery-sales-forecasting,m5-forecasting-accuracy}, and one cold-start electricity consumption forecasting scenario, based on the UCI Electricity dataset \citep{individual_household_electric_power_consumption_235}.

Favorita contains daily sales data for products sold across multiple stores of a large Ecuadorian retail chain. We restrict it to a single store, so that each product maps to exactly one series. M5 contains daily sales data for products sold across multiple US states. We similarly restrict it to a single state (California) and aggregate across that state's stores, so that each product again maps to exactly one series. The UCI Electricity dataset contains hourly consumption readings for 370 clients, each of which already constitutes a single series. 

For each dataset we randomly create disjoint meta-training, meta-validation, and meta-test sets. Meta-validation series are one forecast horizon longer than meta-training series, and meta-test series are one forecast horizon longer than meta-validation series, with the final $H$-length window serving as its scored query horizon. During validation and evaluation, the available history of each validation and test series is limited to a maximum lookback of $L=K \times H$, where $H$ is the forecast horizon and $K$ the number of support windows evaluated. We deliberately keep the meta-training set small with the goal of emulating a low-data regime, in which a global model struggles with generalization to unseen series. We also apply no per-entity normalization, since cold-start series, by construction, offer too little history to estimate reliable per-entity statistics. Dataset characteristics, and how inputs and targets are instead transformed and scaled, are summarized in Appendix~\ref{sec:appendix_datasets}.

\subsection{Naive Baseline}
\label{sec:naivebaseline}
Alongside the base TFT and TMAML-TFT, we report an empirical-quantile naive baseline at every $K$, given the same information budget as the deep models: only the $K$ preceding support windows of length $H$. At every $K$, the naive baseline predicts, at each quantile level $\tau \in \mathcal{Q}$, the empirical $\tau$-quantile of whatever history is available, repeated across every step of the forecast horizon $H$. At $K\geq1$, this history is the pooled target values from the $K$ support windows preceding the query. At $K=0$, no series-specific history exists yet, so the naive baseline instead uses the pooled (transformed) target values from the full meta-training set. This gives a genuine predictive distribution rather than a flat point forecast, making WQL and MACE fair to compare against it, and keeps it information-matched to the deep models throughout.

\subsection{Training Procedure}
\label{sec:trainingparams}
For each dataset and each $K$ studied, we train one TFT using standard ERM and one TMAML-TFT. TFT architecture and data seen at a given $K$ are identical for both procedures. Both procedures draw from the exact same $(K{+}1)H$-length meta-windows. ERM treats this span as one ordinary supervised sample: the first $KH$ steps are the encoder input, the final $H$ steps are the decoder target, and a single forward and backward pass optimizes the quantile loss on that target directly. TMAML is handed the identical span, but partitions it into the support and query samples as detailed in Section~\ref{sec:metatask}: it first takes $M$ inner-loop gradient steps on the support loss, then scores the same, unmodified query horizon with the resulting adapted parameters. So at a given $K$, the two procedures are trained on identical windows of the same observed series, and consequently differ only in how they are trained. Full architecture and optimization hyperparameters are given in Appendix~\ref{sec:appendix_training}, Table~\ref{tab:training_params}.

\subsection{Evaluation Procedure}
\label{sec:evaluation}
We report two metrics: the weighted quantile loss (WQL), which scores the accuracy of the full predicted distribution, and the Mean Absolute Calibration Error (MACE), which scores how trustworthy that distribution is, independent of its accuracy. Both are computed identically for the base TFT, TMAML-TFT, and the naive baseline on the meta-test split.

For a held-out meta-test series $i$ evaluated at a given $K$, the \emph{final} score is computed after the model has been adapted on its support sample drawn from the $K$-support window: the resulting predictor outputs one value per quantile level $\tau \in \mathcal{Q} = \{0.1, 0.2, \ldots, 0.9\}$ for every step of the query horizon, $\hat y_{i,t}^{(\tau)}$ for $t \in [t_0, t_0{+}H)$, which we compare against the true, unseen query values $y_{i,t}$.

\textbf{WQL.} At each quantile level $\tau$, the quantile (pinball) loss $\text{QL}_\tau(q,y) = \tau(y-q)$ if $y > q$, else $(1-\tau)(q-y)$, is summed over the query horizon and normalized by the sum of absolute true query values,
\begin{equation}
\text{WQL}_{i,\tau} = \frac{2\sum_{t} \text{QL}_\tau\!\left(\hat y_{i,t}^{(\tau)}, y_{i,t}\right)}{\sum_{t} \left| y_{i,t} \right|}, \qquad
\text{WQL}_i = \frac{1}{9}\sum_{\tau\in\mathcal{Q}} \text{WQL}_{i,\tau}.
\end{equation}

\textbf{MACE.} For each quantile level $\tau$, the empirical coverage $C_\tau$ is the fraction of (series, query-step) pairs whose true value falls at or below the predicted $\tau$-quantile, pooled across all evaluated series and query steps. A well-calibrated predictor has $C_\tau \approx \tau$. MACE is the mean absolute deviation from this ideal,
\begin{equation}
C_\tau = \frac{1}{NH}\sum_{i,t} \mathbb{1}\!\left[y_{i,t} \leq \hat y_{i,t}^{(\tau)}\right], \qquad
\text{MACE} = \frac{1}{9}\sum_{\tau\in\mathcal{Q}} \left| C_\tau - \tau \right|.
\end{equation}

\textbf{Aggregation.} WQL is computed per series and averaged equally across the $N$ evaluated meta-test series, $\text{WQL} = \frac{1}{N}\sum_i \text{WQL}_i$, rather than pooled into a single ratio over the whole test set as in the original formulation \citep{salinas2020deepar,lim2021temporal}. This weights every cold-start series equally. MACE, in contrast, is inherently a pooled statistic. Optimizer, learning rate, and step count for the adaptation procedure are given in Appendix~\ref{sec:appendix_training}, Table~\ref{tab:training_params}.

\section{Results and Discussion}
\label{sec:results}

\begin{figure}[H]
    \centering
    \includegraphics[width=1\linewidth]{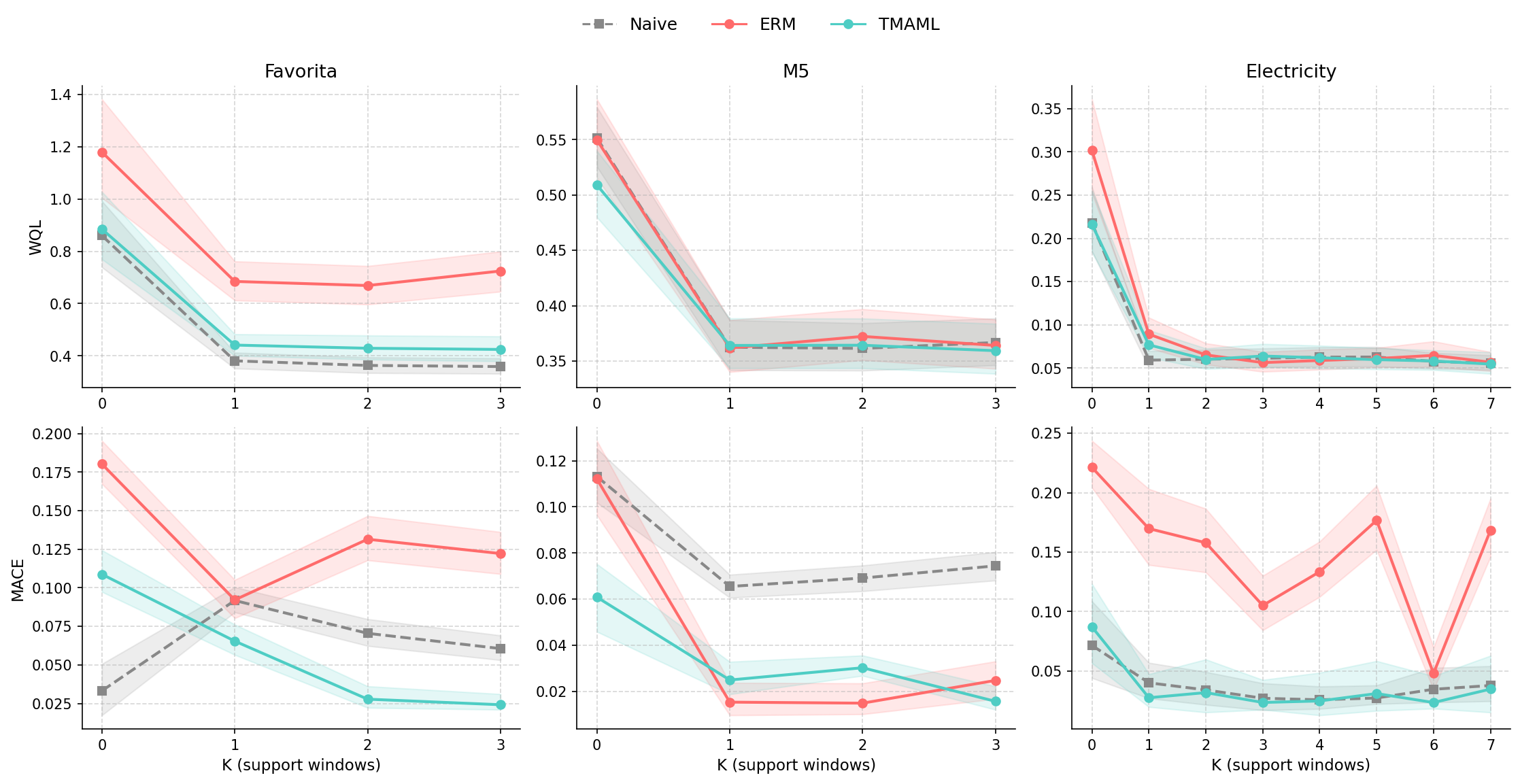}
    \caption{WQL (top) and MACE (bottom) vs.\ $K$, meta-test, for Favorita, M5, and Electricity, left to right. Shaded bands are 95\% bootstrap CIs.}
    \label{fig:results_grid}
\end{figure}

All WQL comparisons below are backed by paired-bootstrap significance tests (1000 resamples, 95\% CI on the paired per-series difference). We report a win only when that interval excludes zero, and a tie otherwise. Figure~\ref{fig:results_grid} shows WQL and MACE as a function of $K$ for each dataset.

\textbf{Favorita.} TMAML beats ERM by a wide, significant margin at every $K$ (25-41\% lower WQL). Naive, however, has a real WQL advantage at $K{=}1,2,3$. At $K{=}0$, before any adaptation happens, TMAML and naive are tied on WQL. Calibration partly reverses the naive picture: naive is best-calibrated only at $K{=}0$. From $K{\geq}1$ TMAML is clearly the best-calibrated model, and its calibration keeps improving as $K$ grows (MACE $0.065\to0.028\to0.024$).

\textbf{M5.} TMAML beats ERM significantly on WQL at $K{=}0,2,3$ and ties at $K{=}1$. It also significantly beats naive on WQL at $K{=}0$ (7.4\% lower) and $K{=}3$ (1.3\%), and is tied with it at $K{=}1,2$. On calibration, TMAML is significantly ahead of ERM at $K{=}0$, ERM is significantly ahead at $K{=}2$, and the two tie at $K{=}1,3$, though naive is calibration-worst at every $K$ without exception.

\textbf{Electricity.} TMAML significantly outperforms ERM on WQL at $K{=}0,1,2,6$, ties at $K{=}4,5,7$, and loses only at $K{=}3$. Naive's WQL edge over TMAML is only significant at $K{=}1$, every other $K$ is a tie. Against ERM, TMAML is clearly best-calibrated, with non-overlapping bootstrap CIs on MACE. Against naive, MACE differences are small and CIs overlap at every $K$.

TMAML consistently beats ERM on WQL (11 of 16 $K$-comparisons significant, only 1 loss). Its comparison with naive is more mixed: naive holds a significant WQL edge on Favorita ($K{=}1,2,3$) and Electricity ($K{=}1$), while TMAML holds the edge on M5 ($K{=}0,3$), with the remaining ten comparisons a tie. TMAML is better-calibrated than ERM on Favorita and Electricity, but only intermittently on M5, where the two trade off by $K$. It is better-calibrated than naive on two of three (Favorita, M5), and statistically indistinguishable from naive on Electricity.

Several factors likely contribute to these results. Meta-training populations differ in transferable structure: Favorita's heterogeneous assortment may be too diverse to yield much beyond a series' own history, while M5's aggregated categories and Electricity's universal consumption cycles offer more. Static covariates are limited too: none for Electricity, and only a handful of categorical attributes for Favorita and M5, leaving TMAML little relational structure to exploit beyond naive's own local history. In addition, no normalization was applied to the cold-start series, so TMAML must learn to adapt to scale and temporal structure simultaneously, and a regime mismatch between the inner and outer loop may destabilize meta-training, since the inner loop's context window is smaller than the outer loop's query context window.


\section{Conclusion, Limitations and Future Work}
\label{sec:conclusion}

We formalised cold-start time series forecasting as a $K$-window learning problem and introduced TMAML, a temporal adaptation of MAML that meta-learns a global model initialization explicitly optimized for adapting from short histories. In a preliminary study, across three cold-start scenarios spanning retail demand and electricity consumption, TMAML-TFT improves on a standard ERM-trained TFT in both probabilistic forecast accuracy (WQL) and calibration, and is better-calibrated than an information-matched naive baseline on two of three datasets, though it does not consistently match naive on WQL itself.

\textbf{Limitations.} Our three scenarios are constructed by truncating the history of series already present in existing benchmark datasets, not from genuinely new entities. Real cold-start series, such as newly launched products, may exhibit different early-life dynamics. Static covariates were limited across datasets, leaving little relational structure for TMAML to exploit. We also did not normalize the cold-start series, so TMAML must learn to adapt to scale together with temporal structure rather than temporal structure alone. Relatedly, the inner loop's context window is smaller than the outer loop's query context window, a regime mismatch that may destabilize meta-training. Each (dataset, model, $K$) configuration is trained once and scored on a single, final query window per series rather than a rolling-origin evaluation. Finally, our experiments cover only a limited set of scenarios and a single architecture, the TFT.


\textbf{Future work.} Promising directions include studying true cold-start series, exploring normalization schemes suited to the cold-start setting, drawing multiple support samples per series to reduce inner loop adaptation variance, and studying larger meta-training populations. Other avenues include evaluating longer forecast horizons, extending TMAML to architectures beyond the TFT, exploring other data scenarios such as new entities with unstructured auxiliary information, and resolving the inner/outer-loop regime mismatch noted above.


\begin{ack}
This work was supported by Flanders Make under Grant 180C0123.
\end{ack}

\bibliographystyle{unsrtnat}
\bibliography{references}

@book{hyndman2018forecasting,
  title={Forecasting: principles and practice},
  author={Hyndman, Rob J and Athanasopoulos, George},
  year={2018},
  publisher={OTexts}
}

@inproceedings{finn2017model,
  title={Model-agnostic meta-learning for fast adaptation of deep networks},
  author={Finn, Chelsea and Abbeel, Pieter and Levine, Sergey},
  booktitle={International conference on machine learning},
  pages={1126--1135},
  year={2017},
  organization={PMLR}
}

@inproceedings{wu2025meta,
  title={Meta-Learning Framework for Effective Few Shot Time Series Prediction},
  author={Wu, Gang and Cong, Li and Huang, Chengbin and Ju, Ying and Jiang, Jiange and Chen, Chen},
  booktitle={2025 IEEE 5th International Conference on Power, Electronics and Computer Applications (ICPECA)},
  pages={18--22},
  year={2025},
  organization={IEEE}
}

@article{caserta2025intermittent,
  title={Intermittent demand forecasting for spare parts with little historical information},
  author={Caserta, Marco and D’Angelo, Luca},
  journal={Journal of the Operational Research Society},
  volume={76},
  number={2},
  pages={294--309},
  year={2025},
  publisher={Taylor \& Francis}
}

@inproceedings{chauhan2020time,
  title={Time series forecasting for cold-start items by learning from related items using memory networks},
  author={Chauhan, Ayush and Prasad, Archiki and Gupta, Parth and Prashanth Reddy, Amiredddy and Kumar Saini, Shiv},
  booktitle={Companion Proceedings of the Web Conference 2020},
  pages={120--121},
  year={2020}
}

@inproceedings{fatemi2023mitigating,
  title={Mitigating cold-start problem using cold causal demand forecasting model},
  author={Fatemi, Zahra and Huynh, Minh and Zheleva, Elena and Syed, Zamir and Di, Xiaojun},
  booktitle={Temporal Graph Learning Workshop@ NeurIPS 2023},
  year={2023}
}

@article{assimakopoulos2000theta,
  title={The theta model: a decomposition approach to forecasting},
  author={Assimakopoulos, Vassilis and Nikolopoulos, Konstantinos},
  journal={International journal of forecasting},
  volume={16},
  number={4},
  pages={521--530},
  year={2000},
  publisher={Elsevier}
}

@article{rangapuram2018deep,
  title={Deep state space models for time series forecasting},
  author={Rangapuram, Syama Sundar and Seeger, Matthias W and Gasthaus, Jan and Stella, Lorenzo and Wang, Yuyang and Januschowski, Tim},
  journal={Advances in neural information processing systems},
  volume={31},
  year={2018}
}

@article{salinas2020deepar,
  title={DeepAR: Probabilistic forecasting with autoregressive recurrent networks},
  author={Salinas, David and Flunkert, Valentin and Gasthaus, Jan and Januschowski, Tim},
  journal={International journal of forecasting},
  volume={36},
  number={3},
  pages={1181--1191},
  year={2020},
  publisher={Elsevier}
}

@inproceedings{rasul2021autoregressive,
  title={Autoregressive denoising diffusion models for multivariate probabilistic time series forecasting},
  author={Rasul, Kashif and Seward, Calvin and Schuster, Ingmar and Vollgraf, Roland},
  booktitle={International conference on machine learning},
  pages={8857--8868},
  year={2021},
  organization={PMLR}
}

@article{oreshkin2019n,
  title={N-BEATS: Neural basis expansion analysis for interpretable time series forecasting},
  author={Oreshkin, Boris N and Carpov, Dmitri and Chapados, Nicolas and Bengio, Yoshua},
  journal={arXiv preprint arXiv:1905.10437},
  year={2019}
}

@inproceedings{challu2023nhits,
  title={Nhits: Neural hierarchical interpolation for time series forecasting},
  author={Challu, Cristian and Olivares, Kin G and Oreshkin, Boris N and Ramirez, Federico Garza and Canseco, Max Mergenthaler and Dubrawski, Artur},
  booktitle={Proceedings of the AAAI conference on artificial intelligence},
  volume={37},
  pages={6989--6997},
  year={2023}
}

@article{lim2021temporal,
  title={Temporal fusion transformers for interpretable multi-horizon time series forecasting},
  author={Lim, Bryan and Ar{\i}k, Sercan {\"O} and Loeff, Nicolas and Pfister, Tomas},
  journal={International journal of forecasting},
  volume={37},
  number={4},
  pages={1748--1764},
  year={2021},
  publisher={Elsevier}
}

@article{nie2022time,
  title={A Time Series is Worth 64Words: Long-term Forecasting with Transformers},
  author={Nie, Y},
  journal={arXiv preprint arXiv:2211.14730},
  year={2022}
}

@article{wang2024timexer,
  title={Timexer: Empowering transformers for time series forecasting with exogenous variables},
  author={Wang, Yuxuan and Wu, Haixu and Dong, Jiaxiang and Qin, Guo and Zhang, Haoran and Liu, Yong and Qiu, Yunzhong and Wang, Jianmin and Long, Mingsheng},
  journal={Advances in Neural Information Processing Systems},
  volume={37},
  pages={469--498},
  year={2024}
}

@inproceedings{das2024decoder,
  title={A decoder-only foundation model for time-series forecasting},
  author={Das, Abhimanyu and Kong, Weihao and Sen, Rajat and Zhou, Yichen},
  booktitle={Forty-first International Conference on Machine Learning},
  year={2024}
}

@article{ansari2024chronos,
  title={Chronos: Learning the language of time series},
  author={Ansari, Abdul Fatir and Stella, Lorenzo and Turkmen, Caner and Zhang, Xiyuan and Mercado, Pedro and Shen, Huibin and Shchur, Oleksandr and Rangapuram, Syama Sundar and Arango, Sebastian Pineda and Kapoor, Shubham and others},
  journal={arXiv preprint arXiv:2403.07815},
  year={2024}
}

@article{liu2025sundial,
  title={Sundial: A family of highly capable time series foundation models},
  author={Liu, Yong and Qin, Guo and Shi, Zhiyuan and Chen, Zhi and Yang, Caiyin and Huang, Xiangdong and Wang, Jianmin and Long, Mingsheng},
  journal={arXiv preprint arXiv:2502.00816},
  year={2025}
}

@article{auer2025tirex,
  title={TiRex: Zero-Shot Forecasting Across Long and Short Horizons with Enhanced In-Context Learning},
  author={Auer, Andreas and Podest, Patrick and Klotz, Daniel and B{\"o}ck, Sebastian and Klambauer, G{\"u}nter and Hochreiter, Sepp},
  journal={arXiv preprint arXiv:2505.23719},
  year={2025}
}

@article{jiang2024metatrans,
  title={MetaTrans-FSTSF: a transformer-based meta-learning framework for few-shot time series forecasting in flood prediction},
  author={Jiang, Jiange and Chen, Chen and Lackinger, Anna and Li, Huimin and Li, Wan and Pei, Qingqi and Dustdar, Schahram},
  journal={Remote Sensing},
  volume={17},
  number={1},
  pages={77},
  year={2024},
  publisher={MDPI}
}

@article{mo2023few,
  title={Few-shot RUL estimation based on model-agnostic meta-learning},
  author={Mo, Yu and Li, Liang and Huang, Biqing and Li, Xiu},
  journal={Journal of Intelligent Manufacturing},
  volume={34},
  number={5},
  pages={2359--2372},
  year={2023},
  publisher={Springer}
}

@inproceedings{arango2021multimodal,
  title={Multimodal meta-learning for time series regression},
  author={Arango, Sebastian Pineda and Heinrich, Felix and Madhusudhanan, Kiran and Schmidt-Thieme, Lars},
  booktitle={International workshop on advanced analytics and learning on temporal data},
  pages={123--138},
  year={2021},
  organization={Springer}
}

@article{zhou2025context,
  title={Context-driven cold-start Web traffic forecasting},
  author={Zhou, Xin and Wang, Weiqing and Buntine, Wray L and Bergmeir, Christoph},
  journal={World Wide Web},
  volume={28},
  number={5},
  year={2025},
  publisher={Springer}
}

@inproceedings{lee2019melu,
  title={MeLU: Meta-Learned User Preference Estimator for Cold-Start Recommendation},
  author={Lee, Hoyeop and Im, Jinbae and Jang, Seongwon and Cho, Hyunsouk and Chung, Sehee},
  booktitle={Proceedings of the 25th ACM SIGKDD International Conference on Knowledge Discovery \& Data Mining},
  pages={1073--1082},
  year={2019}
}

@misc{m5-forecasting-accuracy,
    author = {Addison Howard and inversion and Spyros Makridakis and vangelis},
    title = {M5 Forecasting - Accuracy},
    year = {2020},
    howpublished = {\url{https://kaggle.com/competitions/m5-forecasting-accuracy}},
    note = {Kaggle}
}

@misc{favorita-grocery-sales-forecasting,
    author = {Corporación Favorita and inversion and Julia Elliott and Mark McDonald},
    title = {Corporación Favorita Grocery Sales Forecasting},
    year = {2017},
    howpublished = {\url{https://kaggle.com/competitions/favorita-grocery-sales-forecasting}},
    note = {Kaggle}
}

@misc{individual_household_electric_power_consumption_235,
  author       = {Hebrail, Georges and Berard, Alice},
  title        = {{Individual Household Electric Power Consumption}},
  year         = {2006},
  howpublished = {UCI Machine Learning Repository},
  note         = {{DOI}: https://doi.org/10.24432/C58K54}
}

\appendix

\section{Problem Definition Illustration}
\label{sec:appendix_problemdef}

\begin{figure}[H]
    \centering
    \includegraphics[width=0.85\linewidth]{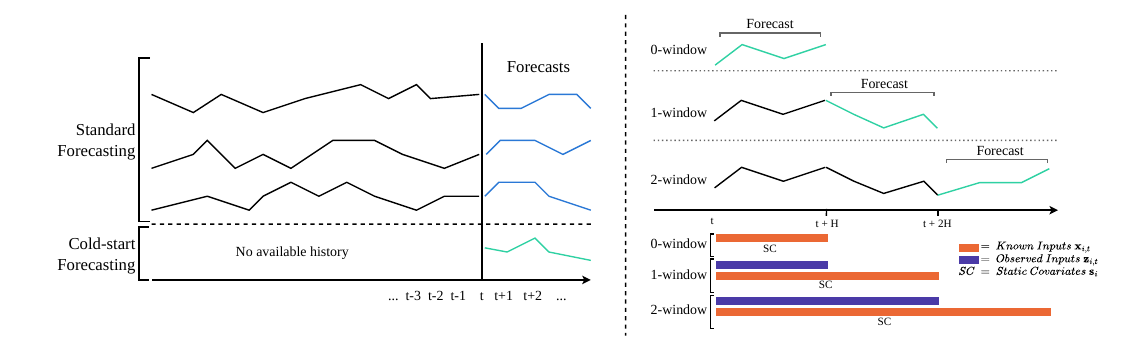}
    \caption{Illustration of the cold-start forecasting problem.}
    \label{fig:coldstart_problem}
\end{figure}

Figure~\ref{fig:coldstart_problem} illustrates the cold-start forecasting problem introduced in Section~\ref{sec:problemdef}. On the left, standard forecasting conditions on a long available history of targets $y$ and observed inputs $\mathbf{z}$, while cold-start entities have no such history available. On the right, our $K$-window formulation, in which known inputs $\mathbf{x}$ and static covariates $\mathbf{s}$ are always available, and an additional span of length $H$ of history in $y$ and $\mathbf{z}$ becomes available at each subsequent window.

\section{TMAML Meta-Training Algorithm}
\label{sec:appendix_algo}

Algorithm~\ref{alg:tmaml} summarizes the full meta-training procedure described in Section~\ref{sec:metatraining}.

\begin{algorithm}[H]
\caption{TMAML Meta-Training}
\label{alg:tmaml}
\begin{algorithmic}[1]
\REQUIRE meta-training series, inner learning rate $\alpha$, outer learning rate $\beta$, inner steps $M$, meta-batch size $B$
\STATE Initialize meta-parameters $\theta$
\WHILE{not converged}
    \STATE Sample a meta-batch of $B$ meta-tasks, at most one per entity
    \FOR{each meta-task (support sample, query sample) in the meta-batch}
        \STATE $\theta_{\text{task}} \leftarrow \theta$
        \FOR{$m = 1$ to $M$}
            \STATE $\theta_{\text{task}} \leftarrow \theta_{\text{task}} - \alpha \nabla_{\theta_{\text{task}}} \mathcal{L}_{\text{support}}(\theta_{\text{task}})$
        \ENDFOR
        \STATE Compute query loss $\mathcal{L}_{\text{query}}(\theta_{\text{task}})$, treating $\theta_{\text{task}}$ as fixed (FOMAML)
    \ENDFOR
    \STATE $\theta \leftarrow \theta - \beta \nabla_{\theta} \frac{1}{B}\sum_{i=1}^{B} \mathcal{L}_{\text{query}}^{(i)}(\theta_{\text{task}}^{(i)})$
\ENDWHILE
\end{algorithmic}
\end{algorithm}

\section{Dataset Construction Details}
\label{sec:appendix_datasets}

\subsection{Splits.} Table~\ref{tab:dataset_stats} summarizes, for each dataset, its temporal frequency, forecast horizon $H$, the largest number of support windows $K$ we evaluate, the corresponding maximum lookback $L=K \times H$, and the number of series in each split. Table~\ref{tab:temporal_splits} gives the calendar date range covered by each split. Splits are defined by disjoint sets of series, and each split's scored horizon lies $H$ steps further into the future than the previous one: meta-validation evaluates one horizon beyond the meta-training period, and meta-test evaluates one horizon beyond that, i.e.\ the meta-validation window shifted forward by exactly $H$. At evaluation time, the history available to each meta-validation and meta-test series is capped at the maximum lookback $L=K\times H$ for the $K$ being evaluated.

\begin{table}[H]
\centering
\caption{Dataset, temporal split, and series-count statistics for the three cold-start scenarios. $H$ is the forecast horizon, $K$ the largest number of support windows evaluated, and $L=K\times H$ the corresponding maximum lookback.}
\label{tab:dataset_stats}
\small
\begin{tabular}{lccc}
\toprule
 & \textbf{Favorita} & \textbf{M5} & \textbf{Electricity} \\
\midrule
\textbf{Freq.}    & Daily & Daily & Hourly \\
\textbf{$H$}      & 28    & 28    & 24 \\
\textbf{Max $K$}  & 3     & 3     & 7 \\
\textbf{Max $L$}  & 84    & 84    & 168 \\
\textbf{Train Series}    & 50    & 50    & 20 \\
\textbf{Validation Series} & 500   & 500   & 85 \\
\textbf{Test Series}     & 1000  & 1000  & 200 \\
\bottomrule
\end{tabular}
\end{table}

\begin{table}[H]
\centering
\caption{Calendar date range covered by each split. The meta-test window is the meta-validation window shifted forward by exactly $H$. Both begin before the meta-training cutoff, though never over the same series (see text).}
\label{tab:temporal_splits}
\scriptsize
\setlength{\tabcolsep}{3pt}
\begin{tabular}{lccc}
\toprule
 & \textbf{Favorita} & \textbf{M5} & \textbf{Electricity} \\
\midrule
\textbf{Meta-train} & 2015-01-02--2015-11-03 & 2011-01-30--2016-02-01 & 2014-01-01--2014-08-07 \\
\textbf{Meta-val}   & 2015-09-09--2015-12-29 & 2015-12-08--2016-03-28 & 2014-08-01--2014-08-08 \\
\textbf{Meta-test}  & 2015-10-07--2016-01-26 & 2016-01-05--2016-04-25 & 2014-08-02--2014-08-09 \\
\bottomrule
\end{tabular}
\end{table}

\subsection{Inputs and target.} Table~\ref{tab:dataset_vars} lists, for each dataset, the known inputs $\mathbf{x}$, observed inputs $\mathbf{z}$, and static covariates $\mathbf{s}$ used, together with the target and any transformation applied to it. Across all three datasets, no per-entity normalization is applied, since cold-start series lack enough history to estimate reliable per-entity statistics. Instead, the target is only log-transformed to compress its scale, and all other real-valued inputs are used as-is (no scaling).

\begin{table}[H]
\centering
\caption{Input variables and target transformation per dataset. All static covariates include the anonymized series identifier in addition to the covariates listed.}
\label{tab:dataset_vars}
\footnotesize
\setlength{\tabcolsep}{3pt}
\begin{tabular}{>{\raggedright\arraybackslash}p{2.3cm}>{\raggedright\arraybackslash}p{2.95cm}>{\raggedright\arraybackslash}p{2.95cm}>{\raggedright\arraybackslash}p{2.95cm}}
\toprule
 & \textbf{Favorita} & \textbf{M5} & \textbf{Electricity} \\
\midrule
\textbf{Known inputs $\mathbf{x}$} & day of month, month, store open flag, on promotion flag, day of week, national, regional, and local holidays & month, year, sell price, weekday, SNAP flag, event name and type & hour, day of week, hours from start \\
\textbf{Observed inputs $\mathbf{z}$} & transactions, oil price & --- & --- \\
\textbf{Static covariates $\mathbf{s}$} & series ID, family, perishable flag, class & series ID, department, category & series ID \\
\textbf{Target (transform)} & sales ($\log y$, floored at $10^{-8}$) & sales ($\log(1{+}y)$) & power usage ($\log(1{+}y)$) \\
\bottomrule
\end{tabular}
\end{table}

\subsection{Series anonymization.} The series identifier is included as a static categorical input, as required by \texttt{pytorch-forecasting}'s \texttt{TimeSeriesDataSet} (its \texttt{group\_ids}), but its embedding is shrunk to a single dimension so the model cannot memorize individual series. This matters because meta-test series carry identifiers never seen during meta-training: a wider embedding would let the model condition on training-series identity, a signal absent at meta-test time, discouraging it from learning the generalizable features cold-start forecasting actually depends on.

\subsection{Licenses.} Favorita and M5 are used under their respective Kaggle competition terms (no separate dataset license is published for either); UCI Electricity is released under CC BY 4.0. We redistribute no raw data from either source.

\section{Training and Meta-Testing Procedure Details}
\label{sec:appendix_training}

Architecture hyperparameters (hidden size, attention heads, dropout) follow the settings reported in the original TFT paper. All other (meta-)hyperparameters in Table~\ref{tab:training_params} were selected via manual exploration on the meta-validation set, for both ERM and TMAML, and the same values are reused for meta-testing adaptation, except where the table notes otherwise.

\subsection{ERM.} Table~\ref{tab:training_params} lists the per-dataset optimizer settings. Electricity's gradient clip value (0.01) and Favorita's (100, effectively no clipping) follow the original TFT paper's settings for those datasets; M5, for which the original paper gives no recommendation, reuses Favorita's value. The learning rate decays on plateau (patience 4 validation checks), \texttt{pytorch-forecasting}'s default TFT scheduler. Training runs for up to 500 epochs with early stopping (minimum improvement $10^{-3}$), validated every 5 training steps against the meta-validation series, with patience 100 checks.

\subsection{TMAML.} Table~\ref{tab:training_params} lists the per-dataset meta-training settings. Gradient clipping is set to 100, effectively no clipping, across all datasets for both the inner and outer loop, found empirically to work best for TMAML. Meta-batches of $B$ tasks are sampled entity-diverse (Section~\ref{sec:metatraining}, i.e.\ at most one task per series per batch). Training runs for up to 150 epochs, validated every 5 steps against the meta-validation series, with early stopping (minimum improvement $10^{-3}$) after 30 checks without improvement in the meta-validation query loss.

\begin{table}[H]
\centering
\caption{Architecture and per-dataset training hyperparameters for ERM and TMAML. $\alpha$/$M$ = inner learning rate / steps, $\beta$ = outer learning rate, $B$ = meta-batch size.}
\label{tab:training_params}
\footnotesize
\setlength{\tabcolsep}{4pt}
\begin{tabular}{lccc}
\toprule
 & \textbf{Favorita} & \textbf{M5} & \textbf{Electricity} \\
\midrule
\multicolumn{4}{l}{\textit{Architecture}} \\
Hidden size          & 240   & 240    & 160 \\
Attention heads      & 4     & 4      & 4 \\
Dropout              & 0.1   & 0.1    & 0.1 \\
Quantile levels      & \multicolumn{3}{c}{0.1--0.9 (step 0.1)} \\
\midrule
\multicolumn{4}{l}{\textit{ERM}} \\
Optimizer            & Adam  & Adam   & Adam \\
LR                   & 0.001 & 0.0001 & 0.001 \\
Batch                & 128   & 128    & 64 \\
Grad.\ clip          & 100   & 100    & 0.01 \\
\midrule
\multicolumn{4}{l}{\textit{TMAML}} \\
Inner optimizer      & SGD   & SGD    & SGD \\
$\alpha$             & 0.001 & 0.001  & 0.001 \\
$M$                  & 10    & 10     & 10 \\
Inner clip           & 100   & 100    & 100 \\
Outer optimizer      & Adam  & Adam   & Adam \\
$\beta$              & 0.0001 & 0.0001 & 0.0001$^\dagger$ \\
$B$                  & 20    & 50     & 20 \\
\midrule
\multicolumn{4}{l}{\textit{Meta-testing adaptation (both ERM and TMAML)}} \\
Optimizer            & \multicolumn{3}{c}{SGD} \\
Steps ($M$)          & \multicolumn{3}{c}{10} \\
LR (ERM / TMAML)      & 0.001 / 0.001 & 0.001 / 0.001 & 0.001 / 0.001 \\
Grad.\ clip          & 100   & 100    & disabled \\
\bottomrule
\end{tabular}

\vspace{2pt}
{\footnotesize $^\dagger$0.001 at $K{=}0$, where the inner loop is inactive.}
\end{table}

\subsection{Compute resources.} All experiments were run on a single NVIDIA RTX 4000 Ada Generation GPU (20GB), with 32 CPU cores and 124GB RAM available. Wall-clock training time varied substantially between the two procedures: TMAML's bi-level optimization, an inner adaptation loop nested within each outer update, makes it inherently more expensive to train than ERM's single-pass optimization. Reproducing the reported results requires on the order of a few GPU-days in total. The broader research process, including exploratory diagnostics and hyperparameter search not reported in the main results, used substantially more compute than this.


\end{document}